\documentclass[11pt,a4paper]{article}
\usepackage[hyperref]{acl2020}
\usepackage{times}
\usepackage{latexsym}
\usepackage{multirow}
\usepackage{graphicx}
\usepackage{enumitem,titlesec,xcolor}
\usepackage[object=vectorian]{pgfornament}
\usepackage{tikz}

\usepackage{microtype}

\usepackage{float}
\floatstyle{boxed} 
\restylefloat{figure}

\aclfinalcopy 

\title{Unmasking Contextual Stereotypes:\\ Measuring and Mitigating BERT's Gender Bias}

\author{Marion Bartl \\
  University of Groningen \\
  University of Malta \\
  \texttt{\small marion.bartl.18@um.edu.mt} \\\And
  Malvina Nissim \\
  University of Groningen \\
  \texttt{m.nissim@rug.nl} \\\And
  Albert Gatt \\
  University of Malta \\
  \texttt{albert.gatt@um.edu.mt} \\}

\date{}

\begin{document}
\maketitle
\begin{abstract}
Contextualized word embeddings have been replacing standard embeddings as the representational knowledge source of choice in NLP systems. Since a variety of biases have previously been found in standard word embeddings, it is crucial to assess biases encoded in their replacements as well. Focusing on BERT \citep{devlin2018bert}, we measure gender bias by studying associations between gender-denoting target words and names of professions in English and German, comparing the findings with real-world workforce statistics. We mitigate bias by fine-tuning BERT on the GAP corpus \citep{webster2018gap}, after applying Counterfactual Data Substitution (CDS) \citep{maudslay2019CDS}. We show that our method of measuring bias is appropriate for languages such as English, but not for languages with a rich morphology and gender-marking, such as German. Our results highlight the importance of investigating bias and mitigation techniques cross-linguistically, especially in view of the current emphasis on large-scale, multilingual language models.
\end{abstract}


\section{Introduction}
The biases present in the large masses of language data that are used to train Natural Language Processing (NLP) models naturally leak into NLP systems. 
These systematic biases can have real-life consequences when such systems are e.g. used to rank the resumes of possible candidates for a vacancy in order to aid the hiring decision \citep{bolukbasi2016man}. If, for example, a model does not associate female terms with engineering professions, because these do not often co-occur in the same context in the training corpus, then the system is likely to rank male candidates for an engineering position higher than equally qualified female candidates.

As NLP applications reach more and more users directly \citep{sun2019litreview}, bias in NLP and as well as resulting societal implications, have become an area of research \citep{hovy-spruit-2016-social,shah2019predictive}. The ACL conference includes a workshop on ethics in NLP since 2017 \citep{2017-acl-gender} and one that specifically addresses gender bias since 2019 \citep{gender-workshop-2019}.

The present work contributes to promoting fairness in NLP by exploring methods to measure and mitigate gender bias in BERT \citep{devlin2018bert}, a contextualized word embedding model.
Its widespread and quick adoption by the research community as the backbone for a variety of tasks calls for an assessment of possible biases encoded in it.

\paragraph{Research Questions} We combine researching how we can measure gender bias in BERT (RQ1) and how such potential gender bias can be mitigated (RQ2), with two further perspectives: a comparison with real-world statistics and a cross-lingual approach. We investigate whether gender bias in BERT is statistically related to actual women's workforce participation (RQ3), and whether a method that we successfully apply to assess gender bias in English is portable to a language with rich morphology and gender marking such as German, since such languages have proven challenging to existing methods \citep{gonen2019gendermarking, zmigrod2019counterfactual, zhou2019examining} (RQ4).

\paragraph{Contributions} This work makes the following contributions:
(i) We present and release the Bias Evaluation Corpus with Professions (BEC-Pro), a template-based corpus in English and German, which we created to measure gender bias with respect to different profession groups. We make the dataset and code for all experiments publicly available at \url{https://github.com/marionbartl/gender-bias-BERT}. 
(ii) Through a more diverse sentence context in our corpus than in previous research, we confirm that the method of querying BERT's underlying MLM (Masked Language Model), proposed by \citet{kurita2019measuring}, can be used for bias detection in contextualized word embeddings. 
(iii) We test our bias analysis on BERT against actual U.S. workforce statistics, which helps us to observe that the BERT language model does not only encode biases that reflect real-world data, but also those that are based on stereotypes.
For bias mitigation, (iv) we show the success of a technique on BERT, which was previously applied on ELMo \citep{elmo_main_paper,zhao2019gender}. Finally, (v) we attempt the cross-lingual transfer of a bias measuring method proposed for English,  
and show how this method is impaired by the morphological marking of gender in German.

\paragraph{Bias Statement}\label{sec:bias_statement}

The present work focuses on gender bias specifically. Gender bias is the systematic unequal treatment on the basis of gender \citep{moss2012science, sun2019litreview}. 
While we are treating gender as binary in this study, we are aware that this does not include people who identify as non-binary, which can create representational harm \citep{blodgett-etal-2020-language}.
In the context of our study of the BERT language model, gender bias occurs when one gender is more closely associated with a profession than another in language use, resulting in biased language models. Before the backdrop of gender participation statistics, we can assess whether a biased representation is related to the employment situation in the real world or based on stereotypes. In the latter case, this constitutes representational harm, because actual participation in the workforce is rendered invisible \citep{blodgett-etal-2020-language}. Moreover, if word representations are used in downstream systems that affect hiring decisions, gender bias, irrespective of whether it is representative of real-world data, may lead to allocational harm, because male and female candidates are not equally associated with a profession from the start \citep{blodgett-etal-2020-language}.


\section{Background and Previous Work}\label{sec:prev_work}

Approaches to gender bias in contextualized word embeddings borrow techniques originally developed for standard embedding models. However, they need to rely on sentence contexts since contextualized word representations are conditioned on the sentence the word occurs in. Previous research uses either templates \citep{may2019measuring, kurita2019measuring} or sentences randomly sampled from a corpus \citep{zhao2019gender, basta2019evaluating}. \citet{may2019measuring} adapt the Word Embedding Association Test \citep[WEAT;][]{caliskan2017semantics} to pooled sentence representations, resulting in the SEAT (Sentence Encoder Association Test). However, the authors express concerns about the validity of this method.
\citet{zhao2019gender} analyze the gender subspace following \citet{bolukbasi2016man}, and also classify the vectors of occupation words that occur in the same context with male and female pronouns, using coreference resolution as an extrinsic measure of gender bias. They mitigate bias via Counterfactual Data Augmentation (CDA) and neutralization.\footnote{Neutralization means that at test time, gender-swapping is applied to an input sentence, and the ELMo representation for both sentences are averaged \citep{zhao2019gender}.} Results show that CDA was more effective.
\citet{basta2019evaluating} measure gender bias by projection onto the gender direction \citep{bolukbasi2016man} as well as clustering and classification, following \citet{gonen2019lipstick}. 
Results from these adapted methods show that contextualized embeddings encode biases just like standard word embeddings \citep{zhao2019gender, basta2019evaluating}.

Instead of adapting bias measuring methods from standard word embeddings, \citet{kurita2019measuring} exploit the Masked Language Model (MLM), native to BERT \cite{devlin2018bert}. Unlike ELMo \citep{elmo_main_paper} or GPT-2 \citep{radford2019language}, BERT learns contextualized word representations using a masked language modelling objective \citep{devlin2018bert}, making the model bi-directional. Crucially, this makes it possible to obtain the probability of a single token in a sentence.
\citet{kurita2019measuring} use the MLM to estimate the probability of a masked, gendered target word being associated with an attribute word in a sentence. This method was shown to capture differences in association across the categories covered by \citet{caliskan2017semantics} in an interpretable way.

One of the problems of current research in NLP is that most work focuses on English \citep{hovy2016social,sun2019litreview}. 
Methods developed to study gender bias in English do not translate well to languages that have grammatical gender, since grammatical gender can have a veiling effect on the semantics of a word. For example, \citet{gonen2019gendermarking} find that words with the same grammatical gender were regarded as more similar, as opposed to words that have similar meanings. This can occur, for instance, because gender agreement between articles and adjectives \citep{corbett_nr_genders30} renders the contexts in which nouns with the same grammatical gender occur more similar.
\citet{gonen2019gendermarking} also found that due to this grammatical gender bias, the debiasing method of \citet{bolukbasi2016man} was ineffective on Italian and German word embeddings. 
\citet{zmigrod2019counterfactual} propose to use CDA as a debiasing method for gender-marking languages, because it is a pre-processing method and as such independent from the resulting vectors. The researchers measure gender bias extrinsically by using a neural language model, following \citet{lu2018CDA}. 

\paragraph{Present Work}
In the present research, we follow \citet{kurita2019measuring} in measuring gender bias. We apply their method of querying the MLM for a more diverse range of sentence templates from a professional context. Additionally, we base the choice of professions on workforce statistics, in order to compare bias to the real-world situation. For mitigating gender bias, we apply \citeauthor{maudslay2019CDS}'s \citeyearpar{maudslay2019CDS} version of CDA to fine-tuning data for BERT, because it has shown promising results for both mitigating bias in English ELMo \citep{zhao2019gender} and in embeddings of morphologically rich languages \citep{zmigrod2019counterfactual}.


\section{Data}\label{sec:data}

In line with previous research \citep{kurita2019measuring, zhao2019gender, basta2019evaluating}, we measure gender bias in BERT using sentence templates. For this purpose we create the \textbf{Bias Evaluation Corpus with Professions (BEC-Pro)}, containing English and German sentences built from templates (Section~\ref{ssec:becpro}).
We also use two previously existing corpora, which are described in Section~\ref{sec:existing-corpora}.

\subsection{Existing Corpora}
\label{sec:existing-corpora}
The Equity Evaluation Corpus (EEC) was developed by \citet{kiritchenko2018EEC} as a benchmark corpus for testing gender and racial bias in NLP systems in connection with emotions. It contains 8,640 sentences constructed using 11 sentence templates with the variables \textless{}person\textgreater{}, which is instantiated by a male- or female-denoting NP; and \textless{}emotion word\textgreater{}, whose values can be one of the basic emotions. We use this corpus for preliminary bias assessment.\footnote{The templates from the EEC corpus were used in preliminary experiments to test the validity of our method. For the sake of space, and to focus on the association with professions, we do not discuss here the results on the EEC data.} This corpus also inspired the structure of the BEC-Pro, and we borrow from it the person words used in our templates.

The GAP corpus \citep{webster2018gap} was developed as a benchmark for measuring gender bias in coreference resolution systems. It contains 8,908 ambiguous pronoun-name pairs in 4,454 contexts sampled from Wikipedia. An example sentence can be found in Figure~\ref{fig:gap_example_sent}.

\begin{figure}
\footnotesize
The historical Octavia Minor's first husband was Gaius Claudius Marcellus Minor, and she bore him three children, Marcellus, Claudia Marcella Major and [Claudia Marcella Minor]; the [\underline{Octavia}] in Rome is married to a nobleman named Glabius, with whom [\underline{she}] has no children.
    \caption{GAP example sentence}
    \label{fig:gap_example_sent}
\end{figure}
We use this corpus to fine-tune BERT (Section~\ref{sec:fine-tuning}).

\subsection{BEC-Pro}\label{ssec:becpro}
In order to measure bias in BERT, we created a template-based corpus in two languages, English and German. The sentence templates contain a gender-denoting noun phrase, or \textless{}person word\textgreater{}, as well as a \textless{}profession\textgreater{}. 

We obtained 2019 data on gender and race participation for a detailed list of professions from the U.S. \citet{labor_statistics}\footnote{\url{https://www.bls.gov/cps/cpsaat11.htm}}. This overview shows, among others, the percentage of female employees for professions with more than 50,000 employed across the United States.
From the lowest-level subgroup profession terms, we selected three groups of 20 professions each: those with highest female participation (88.3\%-98.7\%), those with lowest female participation (0.7\%-3.3\%), and those with a roughly 50-50 distribution of male and female employees (48.5\%-53.3\%). Profession terms were subsequently shortened to increase the likelihood that they would form part of the BERT vocabulary and make them easier to integrate in templates.

For example, the phrase `Bookkeeping, accounting, and auditing clerks', was shortened to `bookkeeper'. 

To maximize comparability, we translated the shortened English professions into both their masculine and feminine German counterparts, using the online dictionary \textit{dict.cc}\footnote{\url{https://www.dict.cc/}}. Translations were corrected by a native speaker of German. 
Feminine word forms were mostly created using the highly productive suffix \textit{-in}.
We note that feminine forms can have a low frequency, which can influence the probability assigned by the language model.
The full list of German professions alongside their English original and shortened counterparts can be found in Table \ref{A:tab:full_professions} in the Appendix.

Following the template-based approach in the EEC \citep{kiritchenko2018EEC}, we created five sentence templates that include a person word, i.e. a noun phrase that describes a person and carries explicit gender information, and a profession term. These templates are shown in Table~\ref{tab:sentence_patterns}. The sentences were first constructed in English and then translated to German. Person words were taken from the EEC and translated into German.\footnote{The phrases `this girl/this boy' were excluded, because they denote children and are therefore less likely to appear in sentences that refer to a professional context. Even though the word `girl' is often used to refer to grown women, this does not apply to the word `boy' to a similar extent.}

\begin{table*}[!ht]
\footnotesize
\centering
\resizebox{\textwidth}{!}{%
\begin{tabular}{p{.06cm}ll}
\hline
& 
\multicolumn{1}{c}{\textbf{English}} &
\multicolumn{1}{c}{\textbf{German}} \\ \hline
\textbf{1} &
  \textless{}person\textgreater is a \textless{}profession\textgreater{}. &
  \textless{}person\textgreater ist \textless{}profession\textgreater{}. \\
\textbf{2} &
  \textless{}person\textgreater{} works as a \textless{}profession\textgreater{}. &
  \textless{}person\textgreater arbeitet als \textless{}profession\textgreater{}. \\
\textbf{3} &
  \textless{}person\textgreater applied for the position of \textless{}profession\textgreater{}. & \textless{}person\textgreater hat sich auf die Stelle als \textless{}profession\textgreater beworben. \\
\textbf{4} &
  \textless{}person\textgreater{}, the \textless{}profession\textgreater{}, had a good day at work. &
  \textless{}person\textgreater{}, die/der \textless{}profession\textgreater{}, hatte einen guten Arbeitstag.\\
\textbf{5} &
  \textless{}person\textgreater wants to become a \textless{}profession\textgreater{}. &
  \textless{}person\textgreater will \textless{}profession\textgreater werden. \\ \hline
\end{tabular}}
\caption{Sentence patterns for English and German}\label{tab:sentence_patterns}
\end{table*}

For example, in English, template 4 in Table \ref{tab:sentence_patterns} could generate the sentence `[My mother], \underline{the} [firefighter], had a good day at work.' The same German template would then generate the sentence \textit{[Meine Mutter], \underline{die} [Feuerwehrfrau], hatte einen guten Arbeitstag.}

For each language, this led to a combined number of 5,400 sentences (5 sentence templates $\times$ 18 person words $\times$ 20 professions $\times$ 3 profession groups).


\section{Method}\label{sec:method}

\subsection{Technical Specifications and Models}\label{ssec:preprocessing}

We use the Huggingface \texttt{transformers} library \citep{Wolf2019HuggingFacesTS} for \texttt{PyTorch} with a default random seed of 42 for all experiments \citep{adams2017ultimate}.

The model used for bias evaluation and fine-tuning is a pre-trained BERT\textsubscript{BASE} model \citep{devlin2018bert} with a language modelling head on top. For reasons of simplicity, this model will be referred to as \textit{BERT language model} from here on. 

For English, the tokenizer and model are loaded with the standard pre-trained uncased BERT\textsubscript{BASE} model.
Unlike in English, where capitalization for nouns is only relevant for proper names (which we do not use), in German capitalization is an integral part of the orthography \citep{stocker2012grammar}. For German we use the cased model provided by DBMDZ.\footnote{\url{https://github.com/dbmdz/berts}}

\subsection{Masking for Bias Evaluation}
The method for measuring bias used in this work is based on the prediction of masked tokens and moreover relies on masking tokens to create potentially neutral settings to be used as prior. In all our experimental settings, \textit{targets} are person words, and \textit{attributes} are professions.

We apply masking to a sentence in three stages, illustrated in Table~\ref{tab:masking}, and add the different masked versions to the BEC-Pro.
Note that only target words (not determiners) are masked. If an attribute contains more than one token, all tokens of the respective phrase are masked individually.

\begin{table}[h]
\centering
\resizebox{\columnwidth}{!}{%
\begin{tabular}{ll}
\hline
\textbf{original}         & My son is a medical records technician.        \\
\textbf{T masked}    & My \texttt{[MASK]} is a medical records technician. \\
\textbf{A masked} & My son is a \texttt{[MASK]} \texttt{[MASK]} \texttt{[MASK]}.  \\
\textbf{T+A masked} & My \texttt{[MASK]} is a \texttt{[MASK]} \texttt{[MASK]} \texttt{[MASK]}. \\ \hline
\end{tabular}%
}
\caption{Masking example}
\label{tab:masking}
\end{table}

\subsection{Pre-processing}\label{sec:input_processing}
The inputs for both measuring and mitigating gender bias largely go through the same pre-processing steps. For GAP corpus instances, which can contain several sentences, we precede these steps by splitting instances into sentences.

As a first step, the fixed input sequence length is determined as the smallest power of two greater than or equal to the maximum sequence length.
In a second step, the inputs are tokenized by the pre-trained \texttt{BertTokenizer} and padded to the previously determined fixed sequence length.
From the padded and encoded inputs, attention masks are created. 
Attention mask tensors have the same size as the input tensors. For each index of the input tensor, non-pad tokens are marked with a \texttt{1} and pad tokens with a \texttt{0} in the attention mask tensor.

\subsection{Measuring Association Bias}\label{ssec:measure_bias}

Following \citet{kurita2019measuring}, who take inspiration from the WEAT \citep{caliskan2017semantics}, we measure the influence of the attribute (A), which can be a profession or emotion, on the likelihood of the target (T), which denotes a male or female person: $P(T|A)$.
It is assumed that in the BERT language model, the likelihood of a token is influenced by all other tokens in the sentence. Thus, we assume that the target likelihood is different depending on whether or not an attribute is present: $P(T) \neq P(T|A)$.
Moreover, we assume that the likelihoods of male- and female-denoting targets are influenced differently by the same attribute word: $P(T_{female} | A) \neq P(T_{male} | A)$.

Following \citet{kurita2019measuring}, we will go on to call the probability of a target word in connection with an attribute word the \textit{association} of the target with the attribute. 

The sentence templates from the BEC-Pro (Section~\ref{ssec:becpro}), are used to measure the association of target and attribute in a sentence. For measuring the association, we need to obtain the likelihood of the masked target from the BERT language model in two different settings: with the attribute masked (prior probability) and not masked (target probability). 
The prior and target probabilities are obtained by applying the softmax function to the logits that were predicted by the BERT language model for the position of the target in the sentence. This produces a probability distribution over the BERT vocabulary for that position in the sentence. We then obtain the (prior) probability of the respective target word by using its vocabulary index. 
The steps to calculate the association are shown in Figure~\ref{fig:logprobsteps}.

\begin{figure}[t]
\footnotesize
\begin{enumerate}[leftmargin=.5cm]
    \item Take a sentence with a target and attribute word\\
    \textit{``He is a kindergarten teacher.''}
    \item Mask the target word\\
    \textit{``[MASK] is a kindergarten teacher.''}
    \item Obtain the probability of target word in the sentence\\
    $p_{T} = P(he = [MASK] | sent)$
    \item Mask both target and attribute word. In compounds, mask each component separately. \\
    \textit{``[MASK] is a [MASK] [MASK].''}
    \item Obtain the prior probability, i.e. the probability of the target word when the attribute is masked\\
    $p_{prior} = P(he = [MASK] | masked\_sent)$
    \item Calculate the association by dividing the target probability by the prior and take the natural logarithm\\
    $\log \frac{p_{T}}{p_{prior}}$
\end{enumerate}
\caption{Procedure to calculate the log probability score, after \citet{kurita2019measuring}.}\label{fig:logprobsteps}
\end{figure}

For interpretation, a negative association between a target and an attribute means that the probability of the target is lower than the prior probability, i.e. the probability of the target \textit{decreased} through the combination with the attribute. A positive association value means that the probability of the target \textit{increased} through the combination with the attribute, with respect to the prior probability. 
Our hypotheses are summarized in Table~\ref{tab:hypotheses}.

\begin{table*}[]
\centering
\resizebox{\textwidth}{!}{%
\begin{tabular}{lll}
\hline
\multicolumn{1}{c}{\textbf{id}} &
  \multicolumn{1}{c}{\textbf{hypothesis}} &
  \multicolumn{1}{c}{\textbf{expected observation}} \\ \hline
\textbf{H1} &
  \begin{tabular}[c]{@{}l@{}}There is a strong association of female (male) \\ person-denoting noun phrases (NPs) with \\ statistically female (male) professions, which \\ is reduced through fine-tuning.\end{tabular} &
  \begin{tabular}[c]{@{}l@{}}Positive association scores between female (male) \\ NPs and statistically female (male) professions,\\ which decrease after fine-tuning.\end{tabular} \\ \hline
\textbf{H2} &
  \begin{tabular}[c]{@{}l@{}}There is a weak association of female (male) NPs \\ with statistically male (female) professions, which \\ is strengthened through fine-tuning.\end{tabular} &
  \begin{tabular}[c]{@{}l@{}}Negative association scores between female (male) \\ NPs and statistically male (female) professions, \\ which increase after fine-tuning.\end{tabular} \\ \hline
\textbf{H3} &
  \begin{tabular}[c]{@{}l@{}}There is no difference between the associations \\ of female and male person-denoting NPs with \\ statistically gender-balanced professions. \\ Associations do not change much after fine-tuning.\end{tabular} &
  \begin{tabular}[c]{@{}l@{}}Both association scores of female and male NPs \\ have approx. the same value, which is likely located \\ around zero. After fine-tuning, the association score \\ does not deviate much from its original value.\end{tabular} \\ \hline
\end{tabular}%
}
\caption{Hypotheses on associations between targets (person words) and attributes (professions) in the BEC-Pro}
\label{tab:hypotheses}
\end{table*}

\subsection{Bias Mitigation}\label{ssec:bias_mitigation}

It has been shown that one of the more effective strategies for removing bias in traditional word embeddings involves modifying the training data instead of trying to change the resulting vector representation \citep{gonen2019lipstick}. One such strategy is a derivative of CDA \citep{lu2018CDA}, Name-based Counterfactual Data Substitution (CDS) \citep{maudslay2019CDS} in which the gender of words denoting persons in a training corpus is swapped in place in order to counterbalance bias. First names are exchanged as well.

To apply CDS in the context of English BERT, we use \citeauthor{maudslay2019CDS}'s \citeyearpar{maudslay2019CDS} code for applying CDS to the GAP corpus \citep{webster2018gap}. Subsequently, these gender-swapped data are used for fine-tuning the English BERT language model. 
Table \ref{tab:hypotheses} illustrates how we expect fine-tuning to influence associations in the English BERT language model. Since GAP instances are balanced between male and female genders, we expect this balance to be preserved after CDS, which would in turn influence male and female entities in the English BERT model to the same extent during fine-tuning. 

\subsection{Fine-tuning}
\label{sec:fine-tuning}

For fine-tuning, each instance in the gender-swapped GAP corpus is tokenized into sentences. Subsequently, the sentences are pre-processed and attention masks are created.
For training, the inputs need to undergo a masking procedure in order to be compatible with BERT's MLM. We follow the standard procedure for masking the inputs, as outlined by \citet{devlin2018bert}.
The masking is carried out using the \texttt{mask\_tokens} function from code by \citet{dontstoppretraining2020}.\footnote{\url{https://github.com/allenai/dont-stop-pretraining/blob/master/scripts/mlm_study.py}} The unchanged input sentences then function as labels.

For training, the instances are randomly sampled and a batch size of one is used. The model is trained for three epochs using an AdamW optimizer with a learning rate of $5 \times 10^{-5}$ and a linear scheduler with warm-up. The fine-tuned model is subsequently used to carry out the exact same bias evaluation as outlined in Section \ref{ssec:measure_bias}.


\section{Results}\label{sec:results}

Table~\ref{tab:results_EN} displays the mean association scores between targets (person words) and attributes (professions) before and after fine-tuning the English BERT language model on the GAP corpus, to which CDS was applied (\textit{pre-association} vs. \textit{post-association}). 
The difference between these two association scores is used to perform the statistical analysis using the Wilcoxon signed-rank test ($W$) for all three profession groups individually.
The effect size $r$ is calculated following \citet{rosenthal1991applied} and \citet{field2012statistics}.
A positive difference score means that the association has increased after fine-tuning, a negative value indicates a decrease in association after fine-tuning. 

\subsection{Overall results}
\begin{table}[h]
\centering
\renewcommand\arraystretch{1.1}
\resizebox{\columnwidth}{!}{%
\begin{tabular}{ccrrrrc}
\hline
\textbf{} &
  \textbf{} &
  \textbf{pre} &
  \textbf{post} &
  \textbf{diff.} &
  \multicolumn{2}{c}{\textbf{Wilcoxon test}} \\ \hline
\textbf{jobs} &
  \textbf{person} &
  \textit{mean} &
  \textit{mean} &
  \textit{mean} &
  \textit{W} &
  \textit{r} \\ \hline
\multirow{2}{*}{\textbf{B}} &
  f &
  -0.35 &
  0.20 &
  0.55 &
  \multirow{2}{*}{359188} &
  \multirow{2}{*}{-0.47} \\
 &
  m &
  0.05 &
  0.07 &
  0.01 &
   &
   \\ \hline
\multirow{2}{*}{\textbf{F}} &
  f &
  0.50 &
  0.36 &
  -0.14 &
  \multirow{2}{*}{96428} &
  \multirow{2}{*}{-0.32} \\
 &
  m &
  -0.68 &
  -0.14 &
  0.55 &
   &
   \\ \hline
\multirow{2}{*}{\textbf{M}} &
  f &
  -0.83 &
  0.13 &
  0.96 &
  \multirow{2}{*}{395974} &
  \multirow{2}{*}{-0.58} \\
 &
  m &
  0.16 &
  0.21 &
  0.05 &
   &
   \\ \hline
\end{tabular}%
}
\caption{Results for English association scores before (pre) and after fine-tuning (post). For jobs, B=balanced, F=female, M=male. In each row, N=900. All $W$ tests are significant at $p=$2e-16.}
\label{tab:results_EN}
\end{table}

Similar to research by \citet{rudinger2018biascoref}, Table \ref{tab:results_EN} contains pro- and anti-typical settings, which correspond to hypotheses H1 and H2, formulated in Table \ref{tab:hypotheses}. 
In the pro-typical setting (H1), male (female) person words are paired with statistically male (female) profession terms. 
Conversely, in the anti-typical setting (H2), male (female) person words are paired with statistically female (male) profession terms. 
Table \ref{tab:results_EN} shows that in fact, there are positive pre-association values in both pro-typical settings and negative pre-association values in both anti-typical settings, which confirms hypotheses H1 and H2. In other words, bias in BERT corresponds to real-world workforce statistics.

For the balanced professions, we expected that association values would not change much as a result of fine-tuning (H3).
This hypothesis could only be confirmed for the male person words, while the female person words show a negative pre-association (-0.35) that changes to a positive post-association (0.20). This shows that male person words hold a neutral position with respect to gender-balanced professions. For female person terms, however, the negative pre-association shows that the gender-parity in the real world data is not reflected in the English BERT language model. 

In general, male person words are relatively stable in BERT.
Associations for these are less strong, i.e. less affected by the presence of the profession words, and also less affected by fine-tuning. These results correspond to \citeauthor{kurita2019measuring}'s \citeyearpar{kurita2019measuring} finding of strong male bias in BERT. 
Further support for this can be found in the results for the balanced profession group, which show similar behavior to those for the male group, though with lower absolute values. This suggests that workers in non-stereotypical professions are more likely to be talked about with male person terms.

In contrast, female person words have higher positive scores in pro-typical settings and lower negative scores in anti-typical settings, which are more susceptible to change after fine-tuning, resulting in positive scores for all professions after fine-tuning. 
On one hand, the more extreme association scores of female terms, as compared to male terms, illustrate them as more marked in language; on the other hand, it shows that the representations of female person words can be more easily adapted.

\subsection{Profession results -- English}

This section zooms in on each individual profession group. 
The results for all profession groups are presented as two bar graphs, the upper graph showing the pre-associations and the lower showing the post-associations. The individual professions are ordered in descending order by the absolute difference in association before and after fine-tuning. 

\begin{figure*}[t]
    \centering
    \includegraphics[width=0.9\textwidth]{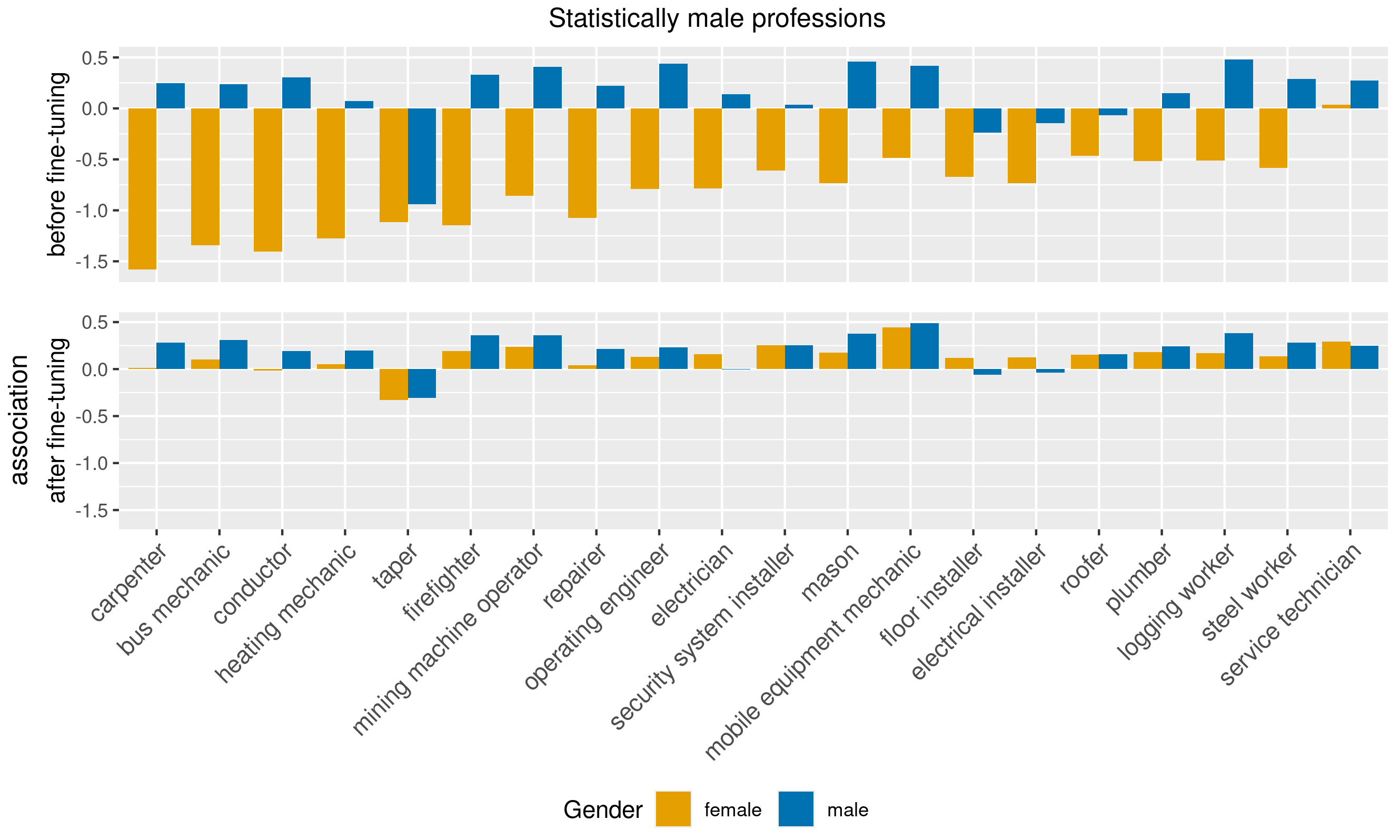}
    \caption{Pre- and post-associations of female and male person words with statistically male professions}
    \label{fig:male_assoc_EN}
\end{figure*}

\paragraph{Male Professions}
Figure \ref{fig:male_assoc_EN} shows the associations before and after fine-tuning for professions with predominantly male workers. It can be seen that there are nearly only negative associations before fine-tuning for female person words with these professions. After fine-tuning, the associations for female person words increase and almost all professions show a positive association with female person words.
The male person words have small positive associations which do not change drastically after fine-tuning, in contrast to female profession terms.
Generally, fine-tuning brings the association values of male and female person words closer, which indicates mitigation of gender bias. The exception to this trend is the word \textit{taper}, whose behaviour can be attributed to the ambiguity of the term, whose more common sense is `narrowing towards a point', rather than the profession.

\paragraph{Female Professions}

\begin{figure*}[h]
    \centering
    \includegraphics[width=0.9\textwidth]{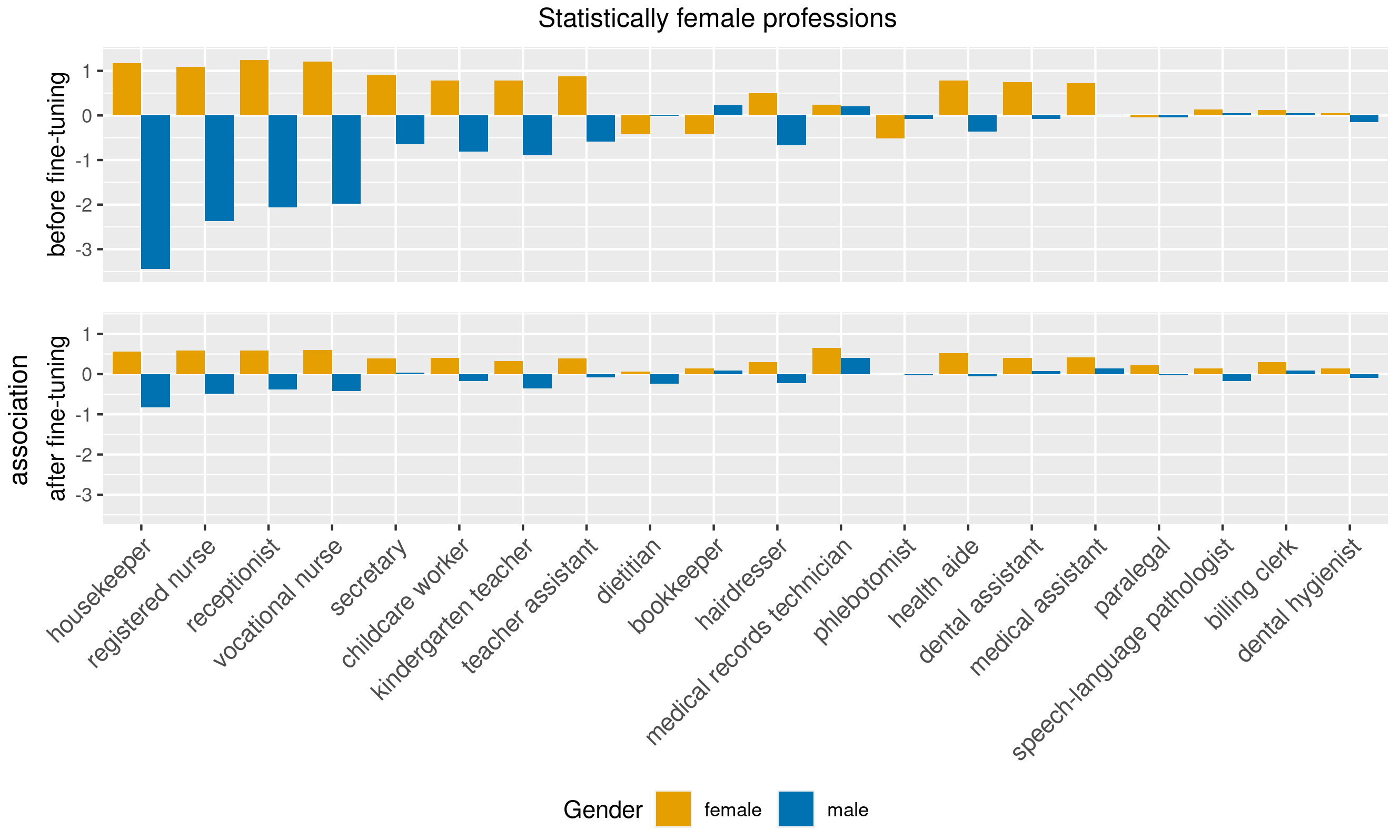}
    \caption{Pre- and post-associations of female and male person words with statistically female professions}
    \label{fig:female_assoc_EN}
\end{figure*}

The results for the statistically female professions are summarized in Figure \ref{fig:female_assoc_EN}. Before fine-tuning, Figure \ref{fig:female_assoc_EN} depicts very strong association values for more stereotypical professions, such as \textit{housekeeper}, \textit{nurse}, \textit{receptionist} or \textit{secretary}. For male person words, these associations are highly negative before fine-tuning and remain negative after. This could be due to the fact that the values were more extreme to begin with. Female person words show positive associations that are less extreme and have a narrower range after fine-tuning.
In contrast, on the far right-hand side of Figure \ref{fig:female_assoc_EN}  (\textit{paralegal}, \textit{speech-language pathologist}, \textit{billing clerk}, \textit{dental hygienist}),  the associations are very low for both female and make person terms, suggesting they are more gender-neutral in English BERT.
Overall, Figure \ref{fig:female_assoc_EN} shows that female bias was reduced, but the model still retained a preference for female person words in context with these professions, which corresponds to the real-world statistics.

\paragraph{Balanced Professions}

\begin{figure*}[t]
    \centering
    \includegraphics[width=0.9\textwidth]{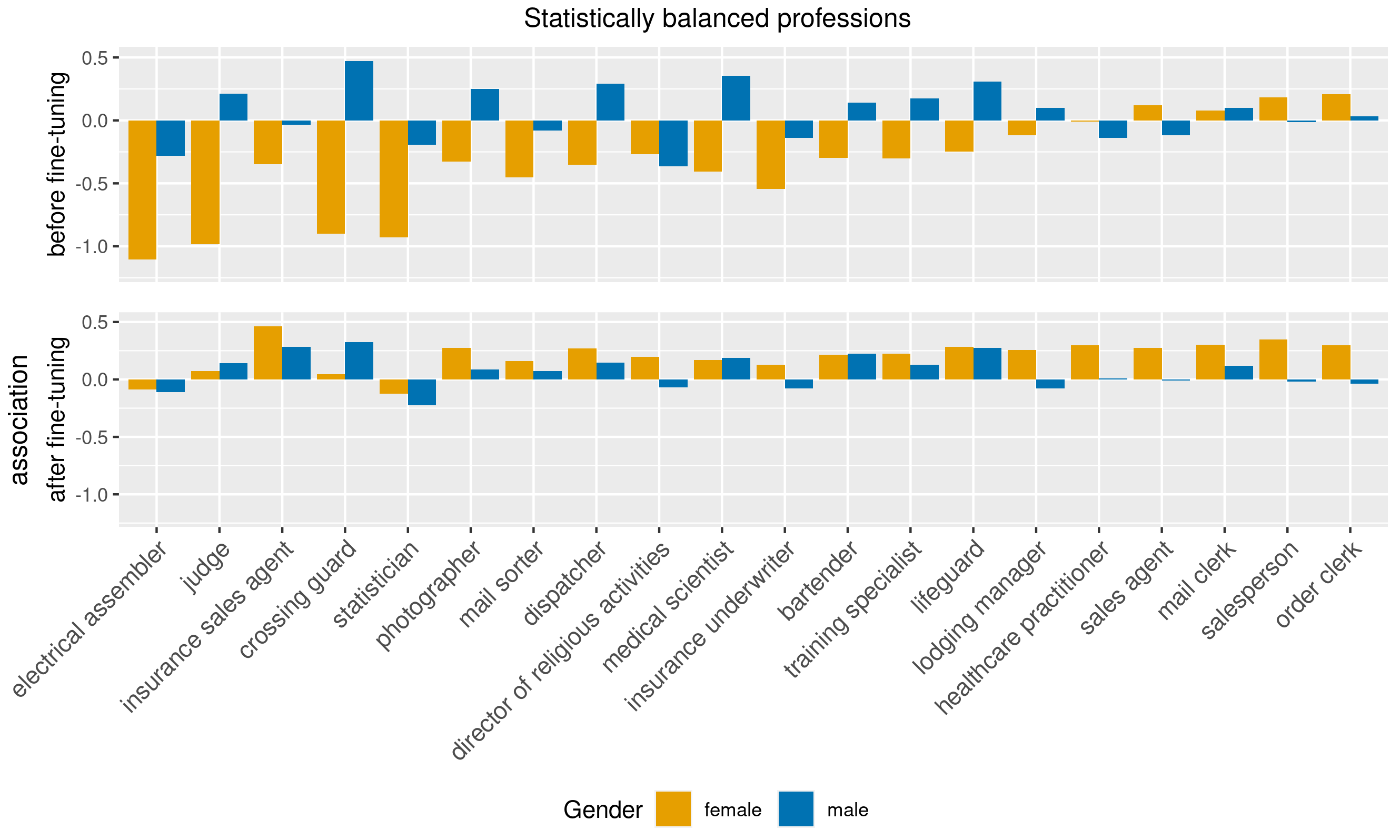}
    \caption{Pre- and post-associations of female and male person words with statistically balanced professions}
    \label{fig:balanced_assoc_EN}
\end{figure*}

The results for the statistically balanced professions are displayed in Figure~\ref{fig:balanced_assoc_EN}. They are especially interesting, because strong male or female biases do not correspond to real-world data and can be ascribed to language use in BERT's training data. 

Figure~\ref{fig:balanced_assoc_EN} shows that the general trend for the associations of female person words with balanced professions before fine-tuning follows the results for statistically male professions: there are mostly negative pre-associations for female person words, which exposes bias in the English BERT language model. These associations mostly become positive after fine-tuning.

For male person words, the results show both negative and positive pre-associations. Professions with negative pre-associations for male person words are generally very specific (such as \textit{electrical assembler} or \textit{director of religious activities}), therefore, the negative associations may be due to low frequency of these terms.
Professions with a positive pre-association for male person words are e.g. \textit{crossing guard}, \textit{medical scientist}, or \textit{lifeguard}. These are more common, therefore, the positive association values reveal male-favoring bias in BERT for the professions in question. Figure~\ref{fig:balanced_assoc_EN} shows converging levels of association after fine-tuning, illustrating the method's effectiveness in mitigating gender bias.

\subsection{Profession Results -- German}\label{ssec:results_german}

Due to the ineffectiveness of the method for German, we only report on pre- and not on post-associations in Table \ref{tab:results_DE}. In order to statistically test the difference between associations for male and female person words, the Wilcoxon signed-rank test was again computed for each profession group separately. 

Table \ref{tab:results_DE} shows that the results across all three profession groups are highly similar: the mean associations for female person words have a value of around 2.1, and the values for male person words are around 1.4. This difference between the groups of person words is significant in all three profession groups with a medium effect size. Nevertheless, the fact that all three groups follow the same pattern indicates that the associations do not capture social gender bias. This can also be observed when looking at the pre-associations for the individual professions (We show them in Figure \ref{A:fig:assoc_DE} in the Appendix).

\begin{table}[]
\centering
\resizebox{\columnwidth}{!}{%
\begin{tabular}{ccccccc}
\hline
\textbf{}            & \textbf{}       & \multicolumn{2}{c}{\textbf{pre association}} & \multicolumn{3}{c}{\textbf{Wilcoxon}} \\ \hline
\textbf{jobs} & \textbf{person}  & \textit{mean}          & \textit{sd}         & \textit{p}  & \textit{W} & \textit{r} \\ \hline
\multirow{2}{*}{\textbf{B}} & f  & 2.14 & 2.4  & \multirow{2}{*}{2e-16} & \multirow{2}{*}{315,058} & \multirow{2}{*}{-0.34} \\
                     & m                  & 1.36                   & 2.06                &             &            &            \\ \hline
\multirow{2}{*}{\textbf{F}}   & f  & 2.05 & 2.45 & \multirow{2}{*}{2e-16} & \multirow{2}{*}{304,635} & \multirow{2}{*}{-0.31} \\
                     & m                    & 1.34                   & 2.09                &             &            &            \\ \hline
\multirow{2}{*}{\textbf{M}}     & f  & 2.14 & 2.46 & \multirow{2}{*}{2e-16} & \multirow{2}{*}{297,605} & \multirow{2}{*}{-0.29} \\
                     & m                    & 1.46                   & 2.15                &             &            &            \\ \hline
\end{tabular}%
}
\caption{Results and statistical evaluation for German associations across professions and person words. For jobs, B=balanced, F=female, M=male. The number of instances for each row is 900.}
\label{tab:results_DE}
\end{table}

The common pattern points to the main difference between the German and English profession terms: German terms are divided into masculine and feminine forms (Section~\ref{ssec:becpro}), because they agree with the grammatical gender of the corresponding person word. We believe that this grammatical difference generates similar association values across the three profession groups.

Specifically, the gender marker of the attribute (profession) influences the likelihood of the target (person word). The fact that the associations for female person words are consistently higher corresponds to the marking of the feminine noun form, e.g. with the suffix \textit{-in}, which is attached to the unmarked masculine form. However, even though it is the unmarked word form, the masculine profession term also carries grammatical gender information, which we assume causes high positive associations across all profession groups.


\section{Discussion and Conclusions}\label{sec:discussion}

The goal of this work is to measure and mitigate gender bias in English and German BERT models \citep{devlin2018bert}.

For measuring gender bias, we use a method first proposed by \citet{kurita2019measuring}: word probabilities taken from the BERT language model are used to calculate association bias between a gender-denoting target word and an attribute word, such as a profession. Our success in making gender bias in the English BERT model visible supports the establishment of the method as a unified metric. 
Moreover, we create the BEC-Pro (Bias Evaluation Corpus with Professions), a template-based corpus set in a professional context, which includes professions from three different statistical groups as well as several male and female person words. With this corpus, which we make available to the community, we contribute to streamlining the visualization of gender bias in other contextualized word embedding models.

For mitigating gender bias, we first apply CDS \citep{maudslay2019CDS} to the GAP corpus \citep{webster2018gap} and then fine-tune the English BERT language model on this corpus. We confirm \citeauthor{zhao2019gender}'s \citeyearpar{zhao2019gender} finding that CDA, or CDS in this case, is useful for mitigating gender bias in the English BERT model. 

Using professions based on workforce statistics allows for a comparison of bias in the BERT language model with real-world data.
We find that the English BERT language model reflects the real-world bias of the male- and female-typical profession groups through positive pro-typical associations and negative anti-typical associations before fine-tuning. After fine-tuning, we observe a reduction in association only for female person words and female-typical professions, but there is an increase in association in both anti-typical settings. This lends support to the effectiveness of our bias mitigation method.

However, we also observe that female person words have higher absolute pre-association values in both the pro- and anti-typical settings, and also show greater changes in post-association.

One possible reason could be BERT's male bias, which has been previously investigated by \citet{kurita2019measuring}. Male person terms seem to have a more stable position in BERT, which could cause their probabilities in the model to not vary much depending on the context and make them less susceptible to change through fine-tuning. 
Another explanation for female terms being more affected by fine-tuning could be that the GAP corpus contained somewhat more female pronouns and nouns, but especially first names, after CDS. Fine-tuning on a corpus with a slight surplus of female person words and first names could have made the likelihood of these terms more sensitive to change. 

In the balanced profession group before fine-tuning, we observe that the BERT language model does not only encode biases that reflect real-world data, but also those that are based on stereotypes.
Despite the fact that all of the balanced professions have an approximately even distribution of male and female employees in the U.S. \citep{labor_statistics}, there is a significantly lower, negative association for female person words before fine-tuning. This signifies that women's visibility in these professions is inhibited, i.e. that women are seen as less likely to carry out such a profession.
In general, the associations in the balanced profession group behave similarly to those in the male-typical profession group. Thus, unless a profession is typically carried out by women, such as the professions \textit{kindergarten teacher} or \textit{nurse}, the default `worker' is culturally seen as male.

Our results moreover show that a method that works well for English is not necessarily transferable to other languages. Since German is a gender-marking language, the agreement between the grammatical gender of the person word and the profession influences the associations. Still, the consistently higher associations of female person words compared to male person words illustrate the linguistic markedness of feminine word forms, as opposed to the default masculine forms, in German. 

Furthermore, the fact that English and German both belong to the Germanic language family \citep{wals} highlights that (genetic) linguistic relatedness does not predict the cross-linguistic success of a method. Especially for a relatively new model such as BERT, developing language-specific methods to assess its limitations is crucial to prevent bias propagation to downstream applications in the language concerned. Our lack of success in transferring the method to German emphasizes the need for more typological variety in NLP research as well as language-specific solutions \citep{sun2019litreview, hovy2016social}.

Naturally, there are a number of limitations of this work. We specifically focus on two, here.
Firstly, we work with only one very specific English BERT model, namely the uncased BERT\textsubscript{BASE}. There are many more contextualized word embedding models besides BERT, such as GPT-2 \citep{radford2019language} or ELMo \citep{elmo_main_paper}. Moreover, there have been various developments and enhancements of the initial BERT model, such as DistilBERT \citep{sanh2019distilbert}, ALBERT \citep{lan2019albert}, or RoBERTa \citep{liu2019roberta}. Therefore, future work could focus on gender bias in a variety of models and investigate whether there are common patterns.

Secondly, the present work extensively relies on choices made by the researchers, due to the template-based method of measuring bias. On one hand, the method is dependent on curated lists of person words and profession terms, which already introduce human bias \citep{sun2019litreview}. We tried to partially counteract this bias by basing the choice of professions on recent labor statistics.
On the other hand, the words in the templates themselves influence the target likelihood, because word representations in BERT are dependent on the entire sentence context \citep{devlin2018bert}. Therefore, future research could include a broader variety of sentences, which could also be randomly sampled.

\clearpage
\section*{Acknowledgments}
This work is based on the first author's master thesis, which was conducted under the ERASMUS Mundus Program Language and Communication Technologies (EMLCT). 
We would like to thank the Center for Information Technology of the University of Groningen for providing access to the Peregrine high performance computing cluster.

\bibliography{anthology,acl2020,extra}
\bibliographystyle{acl_natbib}


\vspace*{1cm}

\hrule

\vspace*{1cm}

\appendix

\section*{Appendix}
\label{sec:appendix}

This appendix contains the detailed list of all professions used in this paper in Table \ref{A:tab:full_professions}. The professions were chosen from a list of professions provided by the U.S. \citet{labor_statistics} based on the percentage of women employed, shortened and subsequently translated into German masculine and feminine forms. Table \ref{A:tab:full_professions} is referred in Section~\ref{ssec:becpro}, where we describe how our template-based BEC-Pro was created.

\begin{figure*}
    \centering
    \includegraphics[height=0.9\textheight]{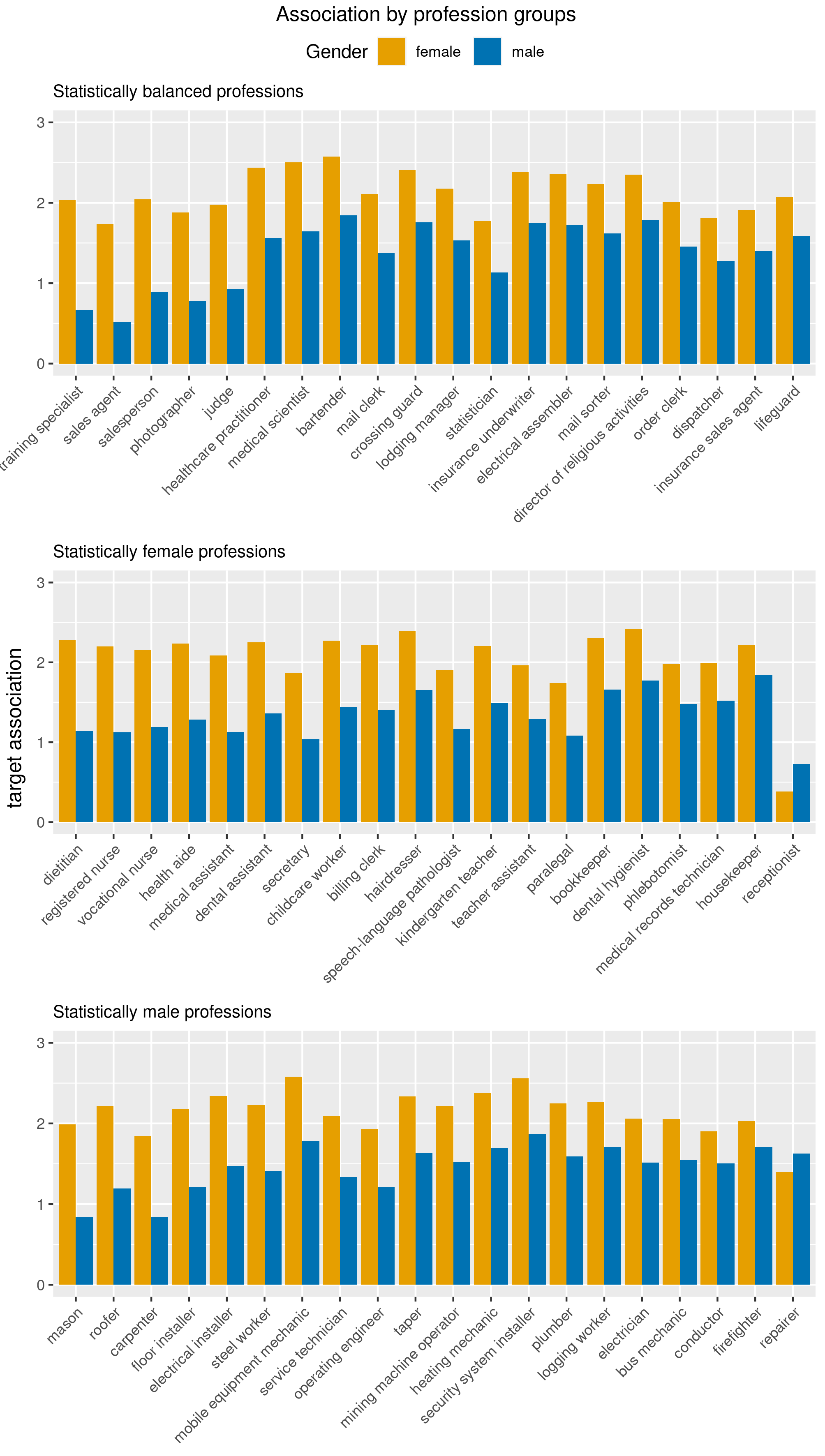}
    \caption{Mean associations for single professions of balanceed, female and male profession groups for German BERT language model}
    \label{A:fig:assoc_DE}
\end{figure*}

Additionally, we provide an illustration of the associations for the German individual professions of the three profession groups in Figure~\ref{A:fig:assoc_DE}. Figure~\ref{A:fig:assoc_DE} is referred in Section~\ref{ssec:results_german} when discussing that the associations do not capture gender bias for the German model due to grammatical agreement. 

\begin{table*}[p]
\centering
\renewcommand\arraystretch{1.1}
\resizebox{\textwidth}{!}{%
\begin{tabular}{llrl}
\hline
\multicolumn{1}{c}{\textbf{original profession}} &
  \multicolumn{1}{c}{\textbf{shortened profession}} &
  \textbf{\begin{tabular}[c]{@{}c@{}}\% \\ women\end{tabular}} &
  \multicolumn{1}{c}{\textbf{\begin{tabular}[c]{@{}c@{}}German profession\\ (masc./fem.)\end{tabular}}} \\ \hline
Preschool and kindergarten teachers &
  kindergarten teacher &
  98.7 &
  Kindergärtner/Kindergärtnerin \\
Dental hygienists &
  dental hygienist &
  96.0 &
  Dentalhygieniker/Dentalhygienikerin \\
Speech-language pathologists &
  speech-language pathologist &
  95.8 &
  Logopäde/Logopädin \\
Dental assistants &
  dental assistant &
  94.9 &
  Zahnarzthelfer/Zahnarzthelferin \\
Childcare workers &
  childcare worker &
  93.4 &
  Kinderbetreuer/Kinderbetreuerin \\
Medical records and health information technicians &
  medical records technician &
  93.3 &
  Medizintechniker/Medizintechnikerin \\
Secretaries and administrative assistants &
  secretary &
  93.2 &
  Sekretär/Sekretärin \\
Medical assistants &
  medical assistant &
  92.7 &
  Arzthelfer/Arzthelferin \\
Hairdressers, hairstylists, and cosmetologists &
  hairdresser &
  92.3 &
  Friseur/Friseurin \\
Dietitians and nutritionists &
  dietitian &
  92.1 &
  Ernährungsberater/Ernährungsberaterin \\
Licensed practical and licensed vocational nurses &
  vocational nurse &
  90.8 &
  Berufskrankenpfleger/Berufskrankenpflegerin \\
Teacher assistants &
  teacher assistant &
  89.7 &
  Betreuungslehrer/Betreuungslehrerin \\
Paralegals and legal assistants &
  paralegal &
  89.6 &
  Rechtsanwaltsgehilfe/Rechtsanwaltsgehilfin \\
Billing and posting clerks &
  billing clerk &
  89.5 &
  Fakturist/Fakturistin \\
Phlebotomists &
  phlebotomist &
  89.3 &
  Phlebologe/Phlebologin \\
Receptionists and information clerks &
  receptionist &
  89.3 &
  Rezeptionist/Rezeptionist \\
Maids and housekeeping cleaners &
  housekeeper &
  89.0 &
  Haushälter/Haushälterin \\
Registered nurses &
  registered nurse &
  88.9 &
  \begin{tabular}[c]{@{}l@{}}staatlich geprüfter Krankenpfleger/\\ staatlich geprüfte Krankenpflegerin\end{tabular} \\
Bookkeeping, accounting, and auditing clerks &
  bookkeeper &
  88.5 &
  Buchhalter/Buchhalterin \\
Nursing, psychiatric, and home health aides &
  health aide &
  88.3 &
  Gesundheitsberater/Gesundheitsberaterin \\ \hline
Drywall installers, ceiling tile installers, and tapers &
  taper &
  0.7 &
  Trockenbaumonteur/Trockenbaumonteurin \\
Structural iron and steel workers &
  steel worker &
  0.9 &
  Stahlarbeiter/Stahlarbeiterin \\
\begin{tabular}[c]{@{}l@{}}Miscellaneous vehicle and mobile equipment \\ mechanics, installers, and repairers\end{tabular} &
  mobile equipment mechanic &
  1.3 &
  \begin{tabular}[c]{@{}l@{}}Mechaniker für mobile Geräte/\\ Mechanikerin für mobile Geräte\end{tabular} \\
Bus and truck mechanics and diesel engine specialists &
  bus mechanic &
  1.5 &
  Busmechaniker/Busmechanikerin \\
\begin{tabular}[c]{@{}l@{}}Heavy vehicle and mobile equipment service technicians \\ and mechanics + Automotive service technicians and mechanics\end{tabular} &
  service technician &
  1.5 &
  \begin{tabular}[c]{@{}l@{}}Kfz-Servicetechniker/\\ Kfz-Servicetechnikerin\end{tabular} \\
\begin{tabular}[c]{@{}l@{}}Heating, air conditioning, and \\ refrigeration mechanics and installers\end{tabular} &
  heating mechanic &
  1.5 &
  \begin{tabular}[c]{@{}l@{}}Heizungsmechaniker/\\ Heizungsmechanikerin\end{tabular} \\
Electrical power-line installers and repairers &
  electrical installer &
  1.6 &
  Elektroinstallateur/Elektroinstallateurin \\
Operating engineers and other construction equipment operators &
  operating engineer &
  1.7 &
  Betriebsingenieur/Betriebsingenieurin \\
Logging workers &
  logging worker &
  1.8 &
  Holzfäller/Holzfällerin \\
Carpet, floor, and tile installers and finishers &
  floor installer &
  1.9 &
  Bodenleger/Bodenlegerin \\
Roofers &
  roofer &
  1.9 &
  Dachedecker/Dachdeckerin \\
Mining machine operators &
  mining machine operator &
  2.0 &
  \begin{tabular}[c]{@{}l@{}}Bergbaumaschinentechniker/\\ Bergbaumaschinentechnikerin\end{tabular} \\
Electricians &
  electrician &
  2.2 &
  Elektriker/Elektrikerin \\
Automotive body and related repairers &
  repairer &
  2.2 &
  Kfz-Mechaniker/Kfz-Mechanikerin \\
Railroad conductors and yardmasters &
  conductor &
  2.4 &
  Schaffner/Schaffnerin \\
Pipelayers, plumbers, pipefitters, and steamfitters &
  plumber &
  2.7 &
  Klempner/Klempnerin \\
Carpenters &
  carpenter &
  2.8 &
  Zimmermann/Zimmerin \\
Security and fire alarm systems installers &
  security system installer &
  2.9 &
  \begin{tabular}[c]{@{}l@{}}Installateur von Sicherheitssystemen/\\ Installateurin von Sicherheitssystemen\end{tabular} \\
Cement masons, concrete finishers, and terrazzo workers &
  mason &
  3.0 &
  Maurer/Maurerin \\
Firefighters &
  firefighter &
  3.3 &
  Feuerwehrmann/Feuerwehrfrau \\ \hline
Retail salespersons &
  salesperson &
  48.5 &
  Verkäufer/Verkäuferin \\
Directors, religious activities and education &
  director of religious activities &
  48.6 &
  \begin{tabular}[c]{@{}l@{}}Leiter religiöser Aktivitäten/\\ Leiterin religiöser Aktivitäten\end{tabular} \\
Crossing guards &
  crossing guard &
  48.6 &
  Verkehrslotse/Verkehrslotsin \\
Photographers &
  photographer &
  49.3 &
  Fotograf/Fotografin \\
\begin{tabular}[c]{@{}l@{}}Lifeguards and other recreational, \\ \\ and all other protective service workers\end{tabular} &
  lifeguard &
  49.4 &
  Bademeister/Bademeisterin \\
Lodging managers &
  lodging manager &
  49.5 &
  Herbergsverwalter/Herbergsverwalterin \\
Other healthcare practitioners and technical occupations &
  healthcare practitioner &
  49.5 &
  Heilpraktiker/Heilpraktikerin \\
Advertising sales agents &
  sales agent &
  49.7 &
  Vertriebsmitarbeiter/Vertriebsmitarbeiterin \\
Mail clerks and mail machine operators, except postal service &
  mail clerk &
  49.8 &
  Postbeamter/Postbeamtin \\
Electrical, electronics, and electromechanical assemblers &
  electrical assembler &
  50.4 &
  Elektro-Monteur/Elektro-Monteurin \\
Insurance sales agents &
  insurance sales agent &
  50.6 &
  Versicherungskaufmann/Versicherungskauffrau \\
Insurance underwriters &
  insurance underwriter &
  51.1 &
  Versicherungsvermittler/Versicherungsvermittlerin \\
Medical scientists &
  medical scientist &
  51.8 &
  \begin{tabular}[c]{@{}l@{}}medizinischer Wissenschaftler/\\ medizinische Wissenschaftlerin\end{tabular} \\
Statisticians &
  statistician &
  52.4 &
  Statistiker/Statistikerin \\
Training and development specialists &
  training specialist &
  52.5 &
  Ausbilder/Ausbilderin \\
Judges, magistrates, and other judicial workers &
  judge &
  52.5 &
  Richter/Richterin \\
Bartenders &
  bartender &
  53.1 &
  Barkeeper/Barkeeperin \\
Dispatchers &
  dispatcher &
  53.1 &
  Fahrdienstleiter/Fahrdienstleiterin \\
Order clerks &
  order clerk &
  53.3 &
  Auftragssachbearbeiter/Auftragssachbearbeiterin \\
\begin{tabular}[c]{@{}l@{}}Postal service mail sorters, processors, \\ and processing machine operators\end{tabular} &
  mail sorter &
  53.3 &
  Postsortierer/Postsortiererin \\ \hline
\end{tabular}%
}
\caption{Shortening and translation of English profession terms into German masculine and feminine forms}
\label{A:tab:full_professions}
\end{table*}

\end{document}